\documentclass[sigconf]{acmart}
\AtBeginDocument{%
  }

\usepackage{multirow}
\usepackage{amsmath}
\usepackage{graphicx}
\usepackage{subfigure}
\copyrightyear{2024}
\acmYear{2024}
\setcopyright{rightsretained}
\acmConference[TAHRI 2024]{2024 International Symposium on Technological Advances in Human-Robot Interaction}{March 9--10, 2024}{Boulder, CO, USA}
\acmBooktitle{2024 International Symposium on Technological Advances in Human-Robot Interaction (TAHRI 2024), March 9--10, 2024, Boulder, CO, USA}
\acmDOI{10.1145/3648536.3648539}
\acmISBN{979-8-4007-1661-4/24/03}

\acmConference[TAHRI 2024]{2024 International Symposium on Technological Advances in Human-Robot Interaction}{March 09--10,
  2024}{Boulder, Colorado}
\acmISBN{978-1-4503-XXXX-X/18/06}

\acmSubmissionID{1028}

\begin{document}

\title{Single-Channel Robot Ego-Speech Filtering during Human-Robot Interaction}


\author{Yue Li}
\email{y6.li@vu.nl}
\orcid{0002-5624-7235}
\affiliation{%
  \institution{Social AI, Vrije Universiteit Amsterdam}
  \streetaddress{De Boelelaan 1111}
  \city{Amsterdam}
  \country{the Netherlands}
  \postcode{1081 HV}
}
\author{Koen Hindriks}
\email{k.v.hindriks@vu.nl}
\orcid{0002-5707-5236}
\affiliation{%
  \institution{Social AI, Vrije Universiteit Amsterdam}
  \streetaddress{De Boelelaan 1111}
  \city{Amsterdam}
  \country{the Netherlands}
  \postcode{1081 HV}
}
\author{Florian Kunneman}
\email{f.a.kunneman@uu.nl}
\orcid{0002-1932-3200}
\affiliation{%
  \institution{Language and Communication, Utrecht University}
  \streetaddress{Heidelberglaan 8}
  \city{Utrecht}
  \country{the Netherlands}
  \postcode{3584 CS}
}
\begin{abstract}
In this paper, we study how well human speech can automatically be filtered when this overlaps with the voice and fan noise of a social robot, Pepper. We ultimately aim for an HRI scenario where the microphone can remain open when the robot is speaking, enabling a more natural turn-taking scheme where the human can interrupt the robot. To respond appropriately, the robot would need to understand what the interlocutor said in the overlapping part of the speech, which can be accomplished by target speech extraction (TSE). 
To investigate how well TSE can be accomplished in the context of the popular social robot Pepper, we set out to manufacture a datase composed of a mixture of recorded speech of Pepper itself, its fan noise (which is close to the microphones), and human speech as recorded by the Pepper microphone, in a room with low reverberation and high reverberation.
Comparing a signal processing approach, with and without post-filtering, and a convolutional recurrent neural network (CRNN) approach to a state-of-the-art speaker identification-based TSE model, we found that the signal processing approach without post-filtering yielded the best performance in terms of Word Error Rate on the overlapping speech signals with low reverberation, while the CRNN approach is more robust for reverberation. Moreover, the best performance is not sufficient for consistent comprehension after filtering, while we see a large diversity in performance across our dataset. We conclude that, first, the human speech volume and pitch strongly affect the performance of the proposed method's results; second, the signal processing method based on speech masking and spectral subtraction is keen to reverberation, while the neural network method is robust; third, the batch normalization layer in TSE models is not useful for filtering the interference speech when it is significantly more powerful than the target speech.
These results show that estimating the human voice in overlapping speech with a robot is possible in real-life application, provided that the room reverberation is low and the human speech has a high volume or high pitch. 
\end{abstract}

\ccsdesc[500]{Human-centered computing~Sound-based input / output}

\keywords{Human-robot interaction, target speech estimation, spectrogram masking, speech recognition}



\received{1 December 2023}

\maketitle

\section{Introduction}
\label{section:1}
Unlike humans who are capable of selective auditory attention \cite{walsh2014selective}, social robots currently cannot prioritize particular sounds. More specifically, they generally lack the ability to extract and recognize human speech when they are talking themselves, since state-of-the-art automatic speech recognition (ASR) systems are not able to separately transcribe such audio streams. Because these systems cannot handle overlapping speech, one approach is to use a simplex channel \cite{skantze2021turn} and configure the robots to listen only to their users when the robot is not talking itself. Rigid and unnatural turn-taking schemes based on this approach are often used, where the microphone needs to be switched off when the robot is talking and switched on again when the robot stops talking. This approach raises several limitations during the human-robot interaction (HRI). For example, during the speech, the robot cannot listen to a user's backchanneling to indicate that they are listening. Furthermore, when the user starts answering the question before the robot finishes, parts of the answers will be left out. This will lead to miscommunication in HRI \cite{skantze2021turn}.

Another approach is to enable a duplex channel, that is, one that allows the system to hear what the user is saying while it is speaking \cite{skantze2021turn}. This approach deploys an additional microphone very close to human users and performs ASR on the signal recorded by this microphone when they start talking \cite{podpora2020human}. In such a setup, the robot voice can be treated as background noise, and ASR systems will most of the time be able to filter it out. A variation on this setup was proposed in \cite{nakadai2000humanoid}. They used two separate (sets of) microphones: one close to the robot speaker and the other relatively far from the speaker and closer to the user. The recordings of one microphone were used as a mask to actively denoise the speaker signal in the other recordings. Neither of these approaches is natural \cite{skantze2009attention}, as they require human users to adjust to a rigid turn-taking scheme or be positioned next to a dedicated microphone.

A more natural approach would be to keep the robot microphone open for the entire duration of the conversation~\cite{schmidt2020acoustic}. This requires the robot to apply a \textit{target speech extraction} (TSE) system\cite{zmolikova2023neural} to the recordings to separate the voice of the user from that of the robot. From an engineering point of view, the TSE problem is directly related to noise reduction and blind source separation (BSS)\cite{zmolikova2023neural,janský2019adaptive}. While regular noise reduction can handle overlapping speech only if the overlapping speaker's voice is known\cite{cohen2009noise}, BSS does not require any information about the target speaker, as is typical for HRI scenarios. However, BSS requires the estimation of the number of speakers and can lead to global permutation ambiguity\footnote{An ambiguous permutation is a permutation which cannot be distinguished from its inverse permutation.}\cite{jain2012blind}. 

Since the success in solving global permutation by deep-clustering \cite{hershey2016deep} and permutation invariant training (PIT) \cite{yu2017permutation}, a large number of neural-based TSE networks have been proposed that are trained and tested on single-channel audio. Jun Du et al.\cite{du2014speech} proposed an initial network based on talker-closed\footnote{TSE is not possible for talkers unseen during training, i.e., not present in the training data.} audio clues to extract the speech of a target talker. Quan Wang et al. \cite{wang2019} proposed a talker-open\footnote{TSE is available for talkers unseen during training, i.e., not present in the training data.} network that extracted a representation of the target talker from a clean enrollment sequence and then isolated the talker's voice in a mix. Similar ideas have also been explored by Meng Ge et al. \cite{ge2020spex} and Katerina Zmolikova et al. \cite{8736286}. 
Other works\cite{ochiai19_interspeech,gu2020multi} use visual and/or spatial clues, which are not the focus of the current study.

\begin{figure}[ht]
  \centering
  \subfigure[Recorded signal.]
    {
        \includegraphics[width=0.43\linewidth]{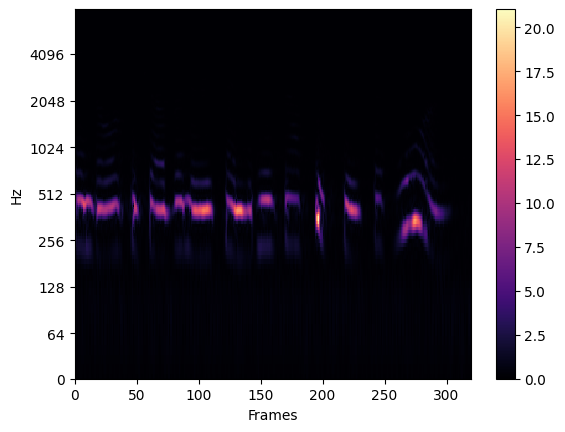}
    }
    \subfigure[TTS generated signal.]
    {
        \includegraphics[width=0.43\linewidth]{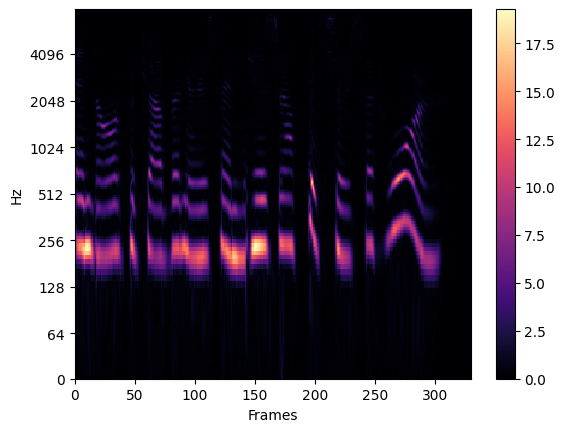}
    }
  \caption{Spectrogram of the recorded speech signal and the corresponding speech signal played by the robot. }
  \label{fig1}
  \Description{The Generated speech signal has more relatively high values on the high frequency bins.}
\end{figure}

Although the results of the research mentioned above are promising \cite{wang2019,janský2019adaptive,ge2020spex}, the gap between laboratory experiments and their deployments in HRI has not yet been demonstrated. This is likely due to three factors.
First, the robot speech signal generated by the \textit{text-to-speech} (TTS) models differs from what its microphone records due to the non-linear and inconsistent microphone response at different frequency levels. As shown in Fig.\ref{fig1}, the speech to be played by the robot speaker has more spectral characteristics than the same speech recorded by its microphones. As a consequence, it is not practical to use the original speech signal to generate a speech mask (SM) or perform spectral subtraction (SS), one of the most widely deployed noise deduction methods, to extract overlapping human speech. Furthermore, this SS-based speech filtering method can over-subtract and result in severe distortion in audio output\cite{wang2019}. It requires post-filtering to restore the authentic target speech. 

\begin{figure}[ht]
    \centering
    \includegraphics[width=0.8\linewidth]
    {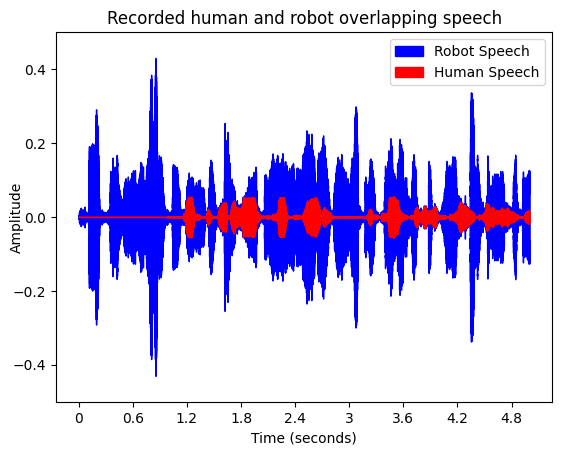}
    \caption{The recorded overlapping speech signal on time domain. }
    \label{fig2}
    \Description{The human speech in the recording is significantly less powerful than the robot speech.}
\end{figure}

Second, the significantly low signal-power ratio between human speech and robot speech also complicates this task, as shown in Fig.\ref{fig2}. In most of the current humanoid robot geometry designs, the close distance between the microphones and speakers causes the robot's speech to possess significantly more power than the overlapping human speech in the received signal. This hinders most speech separation systems in extracting the human speech signal. This is also the reason why this focused task is different from the common TSE task, which could be evaluated on the public benchmark dataset.

Third, most TSE models are deep and computationally intensive, which can lead to unworkable delays for real-time HRI. 

This paper aims to contribute to HRI by enabling a robot to filter out its speech, as well as its stationary ego noise, from the mixture that its microphone receives and improve the speech recognition result of the overlapping human speech signal during HRI. In this work, we address three research questions.
\begin{enumerate}
    \item \textit{Pipeline Design}: How can we filter out robot speech and preserve human speech information from a generated overlapping speech signal?
    \item \textit{Dataset Construction}: How can we construct an overlapping speech dataset that resembles the real recordings?
    \item \textit{Performance Evaluation}: To what extent can performance be improved by post-filtering? What is the trade-off between performance and computational load?
\end{enumerate}

We aim to overcome the gaps mentioned above and enable robots to open microphones during HRI and make sense of what the human says during the overlapping speech. In order to do so, we experiment with methods to remove the speech and ego noise produced by the robot itself from a single-channel recorded audio, and to recover the overlapping human speech signals to improve the ASR result. We propose two different audio processing architectures to filter out the robot's speech, with the help from the robot's embedded API and the online TTS API \cite{googletts} to pre-acquire the interfering speech signal. The experimental results show that the proposed signal processing-based method without post-filtering is most effective in improving the human speech ASR results under the circumstances when the room reverberation is low and the target speaker is high pitched or at a relatively high volume, while the proposed neural network approach shows good robustness to the reverberation condition.

The rest of this paper is organized as follows. In Section \ref{section:2}, we introduce several neural network-based and signal processing-based TSE methods. In Section \ref{section:3}, we present our two proposed pipelines for solving the robot's ego speech filtering problem. In Section \ref{section:4}, we elaborate on the setup of the experiment and the evaluation metrics. In Section \ref{section:5}, we present and analyze the results of the baseline and proposed methods. In Section \ref{section:6}, we draw our conclusions and discuss future work.

\section{Related Work}

\label{section:2}
The field focused on TSE for single-channel recordings, also known as target voice filtering\cite{wang2019}, can be summarized into two approaches: signal processing-based and neural network-based. 
\subsection{Signal Processing-Based TSE}
In the single-channel approach, signal processing-based TSE methods generally calculate and reduce audio noise in a spectrum space. The enhanced signal $\hat{S}_{tf}$ is obtained by multiplying the input signal $X_{tf}$ by non-negative real-value weights, $W_{tf}$\cite{1163209}, also known as the signal mask (SM). Ideally, the SM is 0 if only the undesired signal is active and 1 if the desired signal is active in a certain time-frequency (TF) bin. A wide variety of approaches have been proposed to optimize this SM \cite{benesty2009noise}, including spectral subtraction, the Wiener filter, minimum mean-square estimation, the factorial hidden Markov model, and minima-controlled recursive averaging. These methods have been commonly designed and implemented to estimate SM during speech pause or silence. They are efficient in attenuating stationary noise\cite{wake2019enhancing}. Several spectral subtraction schemes have been proposed for robotics \cite{ince2009ego}. But they were mainly aimed at estimating target speech from robot's ego fan noise or joint noise.

\subsection{Neural Network-Based TSE}
In the mid-2010s, deep neural networks were introduced for the first time to address the TSE problem. Katerina Zmolikova et al. \cite{vzmolikova2019speakerbeam} introduced SpeakerBeam, which explored three different methods to inform the network to modify the behaviour of the acoustic model. Quan Wang et al. proposed VoiceFilter\cite{wake2019enhancing} and its subsequent work, VoiceFilter-Lite \cite{wang2020voicefilter}, as plug-ins before automatic speech recognition. They used a pre-trained speaker diarization network as an additional informant. Shulin He et al. \cite{he2020speakerfilter} followed this idea and proposed SpeakerFilter, which learned the target speaker's information while producing the SM and no longer required a pre-trained speaker identification network. To avoid the adverse effect on performance from different window lengths when analyzing the reference signal and the input mixture signal, Meng Ge et al. \cite{ge2020spex+} proposed a time-domain solution for TSE, which avoided phase estimation in the TF domain. Although the performance of these proposed networks is promising on public datasets\cite{maciejewski2020whamr}, they have not been tested or evaluated in real recordings during HRI, where human speech overlaps with robot speech.

In this study, we compare these two different methods, aiming to filter out the robot's speech in the overlapping audio mixture. Inspired by the work\cite{wake2019enhancing,wang2019} and their promising results in a related task, we select spectral subtraction and convolutional recurrent neural network (CRNN) as our method to achieve this objective.

\section{Methods}
\label{section:3}
\subsection{Problem Formulation}
\begin{figure}[ht]
    \centering
    \includegraphics[width=0.82\linewidth]
    {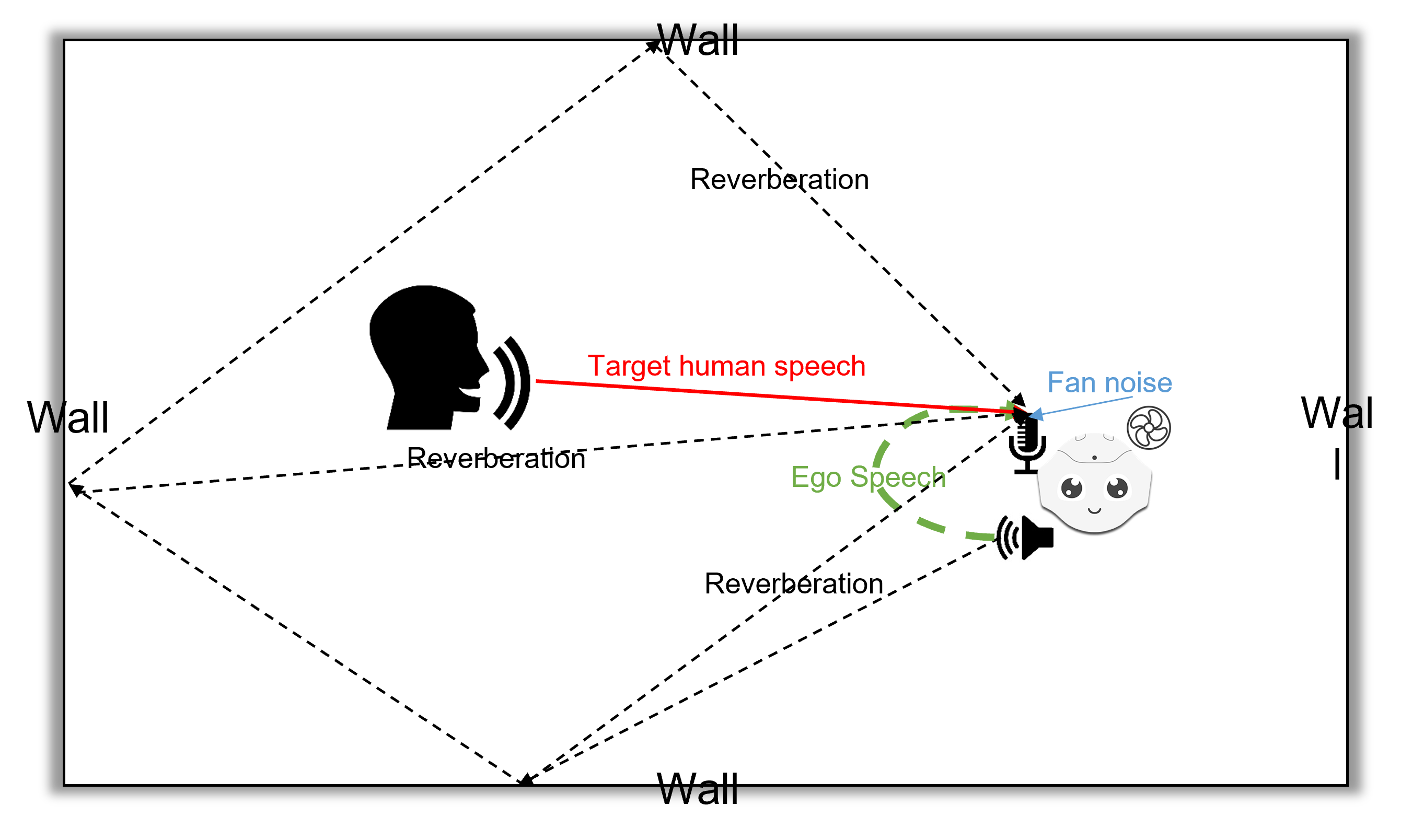}
    \caption{The illustration of the generic application scenario. }
    \label{fig4}
    \Description{There are four kinds of signals in our focused application scenario: robot ego speech, the room reverberation of the ego speech, fan noise, and the target human speech.}
\end{figure}

The generic application can be illustrated in Fig.\ref{fig4}.
In this figure, the observed mixture signal is represented in the short time Fourier Transform (STFT) domain, $X_{tf}$, as,
\begin{equation}
    X(t,f) = S(t,f) + N(t,f)
    \label{eq1}  
\end{equation}
where $S$ is the spectrum of the target speech signal, $N$ is the interfering signal comprising the speech from other speakers and noise (in the experiment we only consider the reverberation from the interfering speaker), and $t$ and $f$ are time and frequency indices, respectively. There is also a reference speech $A$ from the text-to-speech model, which the robot will play and record during the interaction. In this paper, our objective is to extract the target speech $S$ from the mixture signal $X$ with the help of $A$.
\subsection{Proposed Structure}
We propose two different audio processing pipelines to extract human speech when it overlaps with robot speech.

\subsubsection{Audio Processing Pipeline based on Spectral Subtraction}
\begin{figure}[ht]
    \centering
    \includegraphics[width=0.9\linewidth]
    {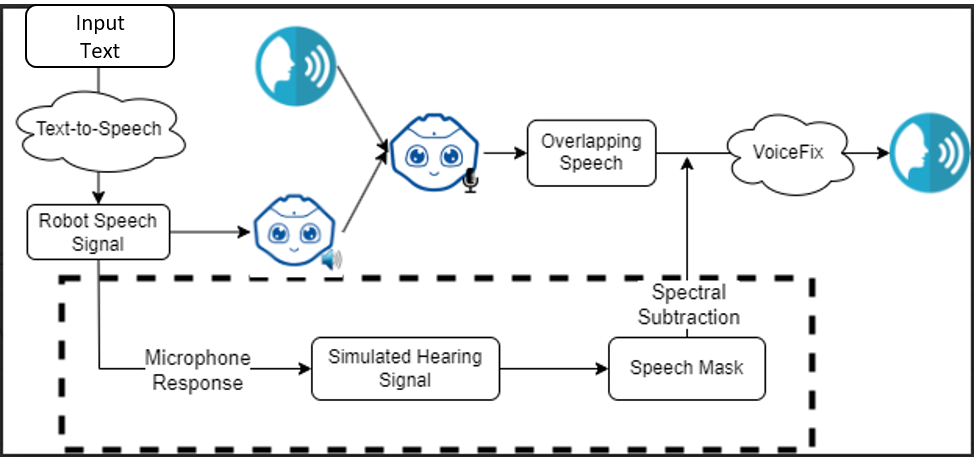}
    \caption{Signal Processing Pipeline. In the pipeline, the input texture is sent to on cloud text-to-speech API using SIC framework. }
    \label{fig3}
    \Description{The generated speech signal is then uploaded to the robot and played. Meanwhile, The correlating speech mask is generated by  microphone response function then used to estimate the target human speech by spectral subtraction. A pre-trained model, VoiceFixer, is avaliable to post-filter and restore the estimated signal.}
\end{figure}

Inspired by proprioception \cite{christensen2007premotor}, which enables humans to subconsciously generate a speech mask to filter their own voice when they start talking, and given that a new deployment\footnote{Social Interaction Cloud (SIC) is a framework that enables the users to design and implement a socially interactive robot prototype on a Pepper humanoid robot. By this framework, users can connect to Google TTS APIs and make the robot speak in other sounds than its embodied voice.} \cite{sic} enables the pre-acquisition of the robot speech signal from the TTS APIs, we propose the self-speech filtering pipeline, as shown in Fig.\ref{fig3}. The dashed box highlights the designed ego-speech filtering pipeline. 

As mentioned in Section \ref{section:1}, the reference speech signal differs from the recorded speech signal, due to the non-linear microphone frequency response function. To obtain the SM for the interference speech in the recordings, the microphone response function must be precalculated. The frequency response of the speaker can be defined as below:
\begin{equation}
    x_s(t) = h_s\ast a(t)
\end{equation}
where $x_s(t)$ is the frequency response of the speaker at time $t$, $h_s$ is the response function of that speaker, $\ast$ means convolutional operation, and $a(t)$ is the robot's input signal.
Correspondingly, the response of the microphone can be defined as follows:
\begin{equation}
\centering
    \begin{split}
        x_m(t) &= h_m\ast [x_s(t)+noise(t)]\\
        &=h_m\ast [h_s\ast a(t)+noise(t)]\\
        &=h_m\ast h_s\ast a(t) + h_m \ast noise(t)
    \end{split}
\end{equation}
where $x_m(t)$ is the frequency response of the microphone, $h_m$ is the frequency response function of that microphone, and $noise(t)$ is the ego noise signal. Using the STFT on both sides, we can obtain the following. 
\begin{equation}
    X_m(t,f) = (H_m(f)\cdot H_s(f))\cdot A(t,f) + N(t,f)
    \label{eq2}
\end{equation}
where $X_m(t,f)$, $A(t,f)$, and $N(t,f)$ respectively denote the spectrogram of the received mixture audio signal, the robot's input signal, and the recorded ego noise signal. Furthermore, due to the time-invariant characteristics of the frequency response coefficients, $H_m(f)\cdot H_s(f)$ represents the speaker-microphone frequency response coefficients and can be calculated as follows:
\begin{equation}
    H_m(f)\cdot H_s(f) = \frac{X_m(f)-N(f)}{A(f)}
\label{eq3}
\end{equation}
From Equation \ref{eq3}, the frequency response function between the robot input signal and the recorded signal can be obtained by recording the robot's ego fan noise, as well as a sine signal whose frequency sweeps over all the possible bins \cite{sinesweep}. The SM based on the reference signal spectrogram $A_{ref}$ can be calculated by the following equation:
\begin{equation}
    SM(t,f) = |X_{mix}(t,f)|<= \alpha \times |A_{ref}(t,f)| \cdot H_m(f)\cdot H_s(f)
\label{eq4}
\end{equation}
where $|X_{mix}|$ and $|A_{ref}|$ are respectively the real-value spectrogram of the overlapping speech signal and the reference speech signal, and $\alpha$ is the over-subtraction factor. Considering the non-linear characteristics of speech \cite{mack2019deep}, we apply a Hanning window on the $SM$ produced by Eq.\ref{eq4} and obtain a new $\hat{SM}$:
\begin{displaymath}
    \hat{SM}(t,f) = H(2\times L+1,2\times I+1) \ast SM(t,f)
\end{displaymath}
where $H(2\times L+1,2\times I+1)$ is the Hanning window, $2\cdot L+1$ is the window dimension in time-frame direction, and $2\cdot I+1$ is in frequency direction. In our experiment, we set $L=3$ and $I=1$, resulting in a window dimension of $(7,3)$. The estimated speech signal can be obtained as follows:
\begin{equation}
    X_{est}(t,f) =  \beta\times sign(X_{mix}) [|X_{mix}(t,f)| \cdot (1-\hat{SM}(t,f))] 
    \label{eq9}
\end{equation}
where $X_{est}$ is the spectrogram of the estimation speech, $sign(\ast)$ represents the element-wise indication of the sign of the original $X_{mix}$, and $\beta$ is amplify coefficient. And finally, the estimated speech signal can be obtained by inverse STFT (iSTFT). The estimated speech signal obtained may be distorted due to oversubtraction. We will then use a pre-trained \textit{VoiceFixer} model\cite{liu2021voicefixer}, which shows great performance in restoring strongly degraded human speech, to reconstruct this.

\begin{figure}[ht]
    \centering
    \subfigure[Reference signal.]
    {
        \includegraphics[width=0.43\linewidth]{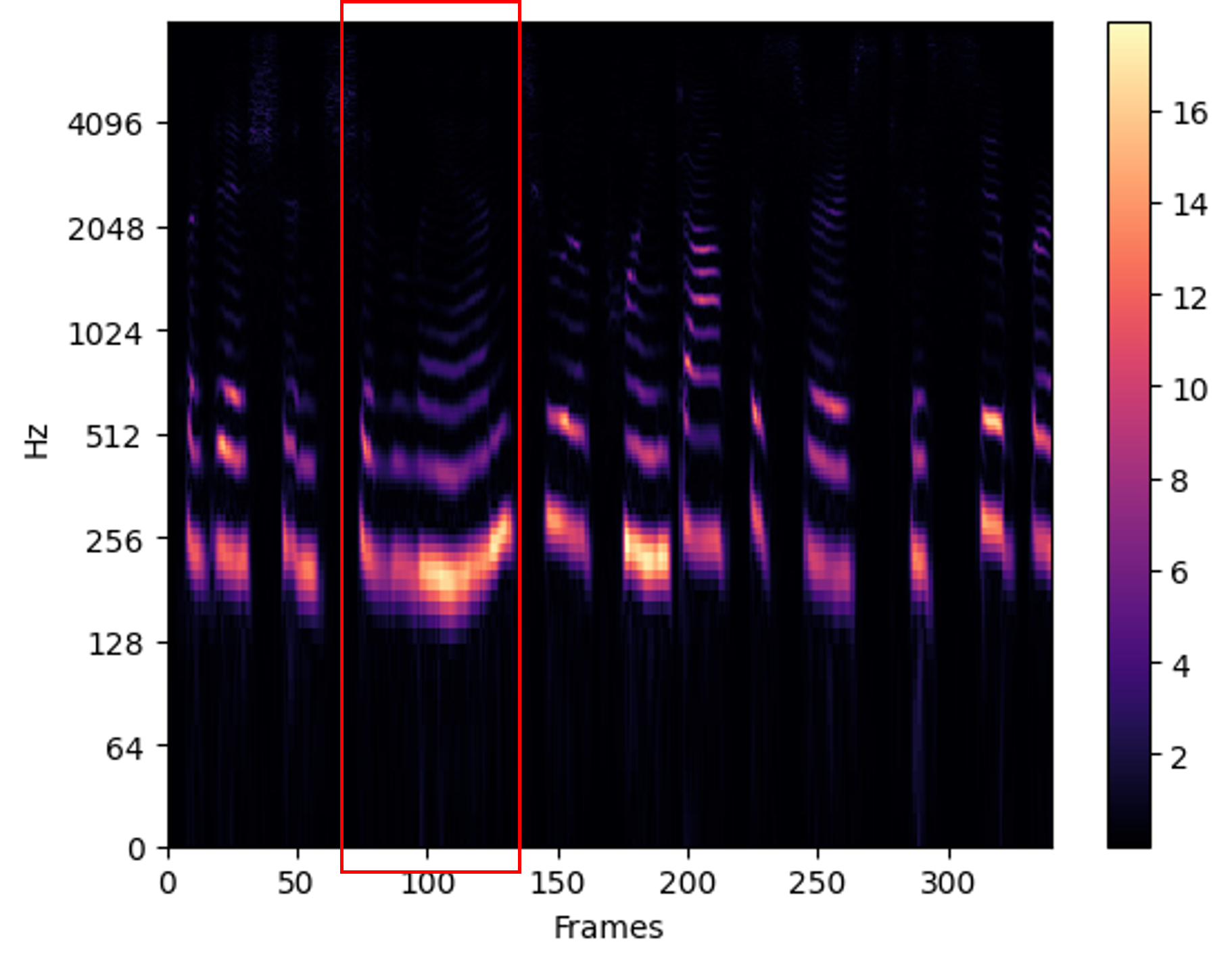}
    }
    \subfigure[Recorded signal after frame power VAD.]
    {
        \includegraphics[width=0.43\linewidth]{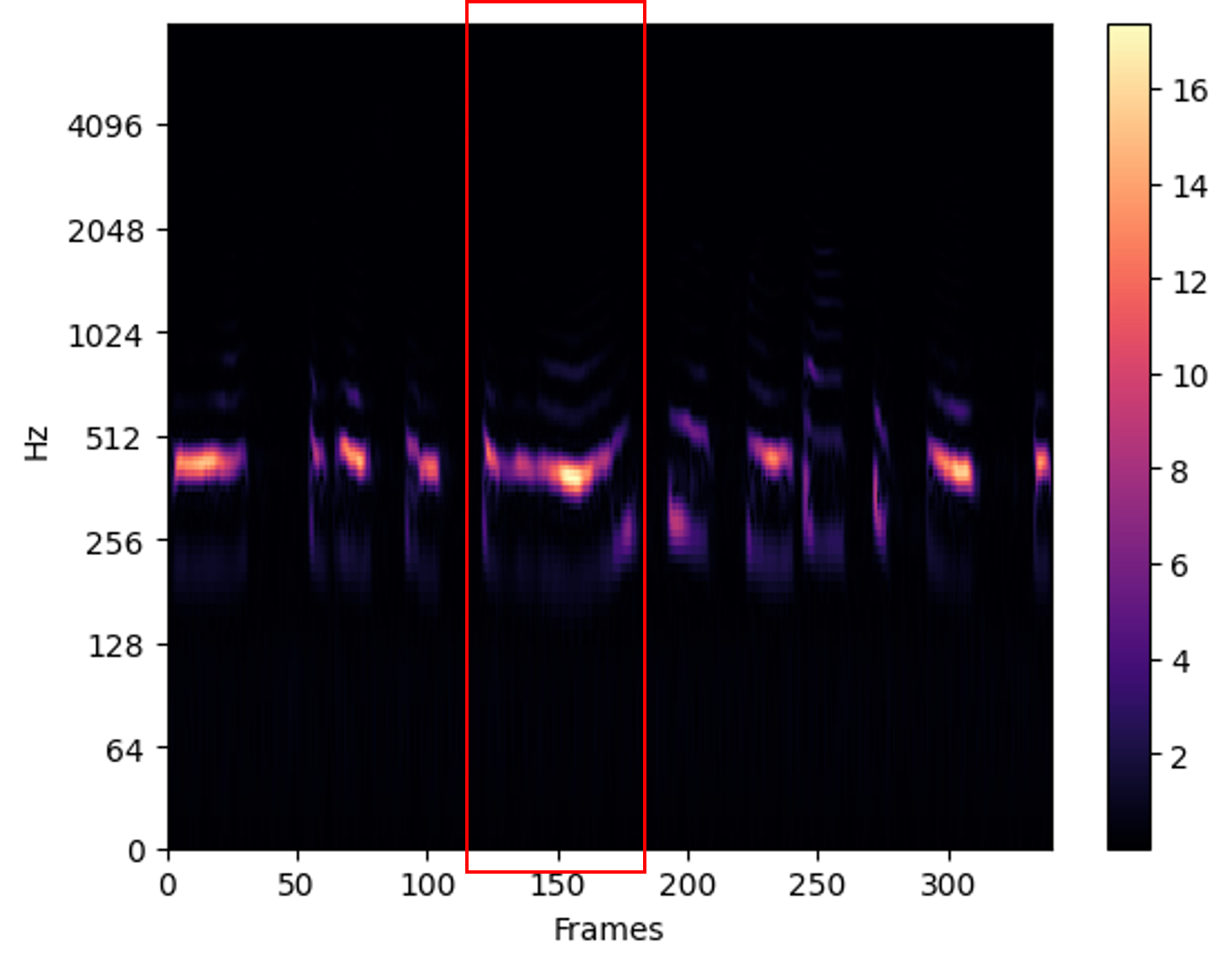}
    }
    \subfigure[Recorded signal after cross-correlation VAD.]
    {
        \includegraphics[width=0.43\linewidth]{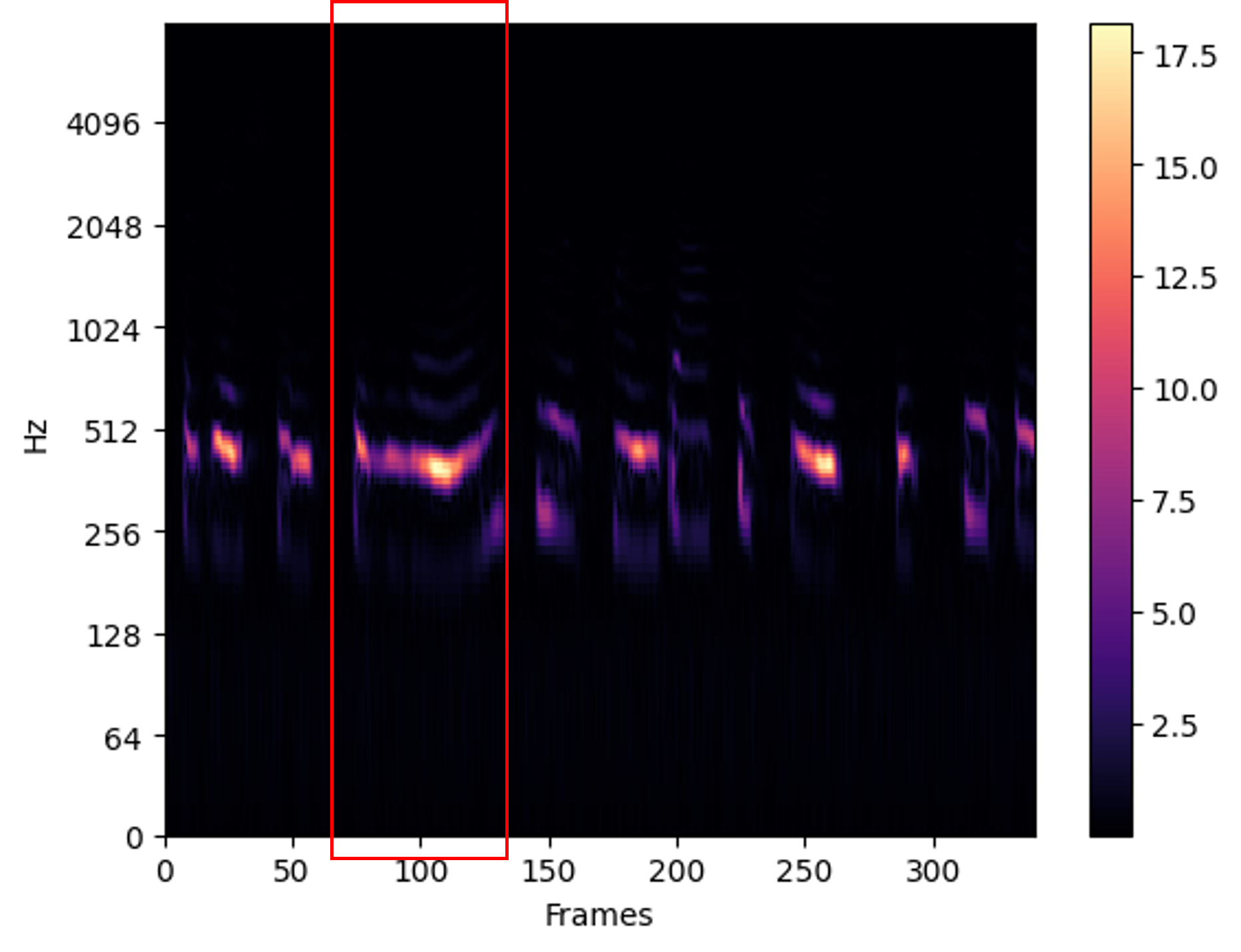}
    }
    \caption{The spectrogram of different signals at time $T_0$. }
    \label{fig7}
    \Description{The silent part in the recorded signal is trimmed based on different methods.  The parts highlighted by red rectangles are the matching frames}
\end{figure}

Another crucial factor in Eq.\ref{eq4} is the matching $t$ between the robot speech and the reference speech. Because we focus on the estimation of human speech when interrupting robot speech and the time delay is unstable when sending the command to make the robot speak during real-life operation, we propose to use the first 0.5 second length of the reference signal as a detector and calculate the cross-correlation (CC) value $cc$ between the detector and the recorded robot speech signal. The maximum value $cc$'s corresponding value $X$ will be considered the time delay between the reference signal and the recorded signal. The comparison result is presented in Fig.\ref{fig7} between the proposed method and the traditional frame power-based voice activity detection (VAD) method\cite{sangwan2002vad}. We can observe that the recorded signal after trimming the silent part based on cross-correlation aligns better with the reference signal in the time-frequency (TF) domain.

In order to alleviate oversubtraction, which is inevitable in spectral subtraction, we tested to adopt a pre-trained model, VoiceFixer \cite{liu2021voicefixer}, to post-filter and restore the estimated signal.

\subsubsection{CRNN Architecture}
\begin{figure*}[ht]
    \centering
    \includegraphics[width=0.85\linewidth]
    {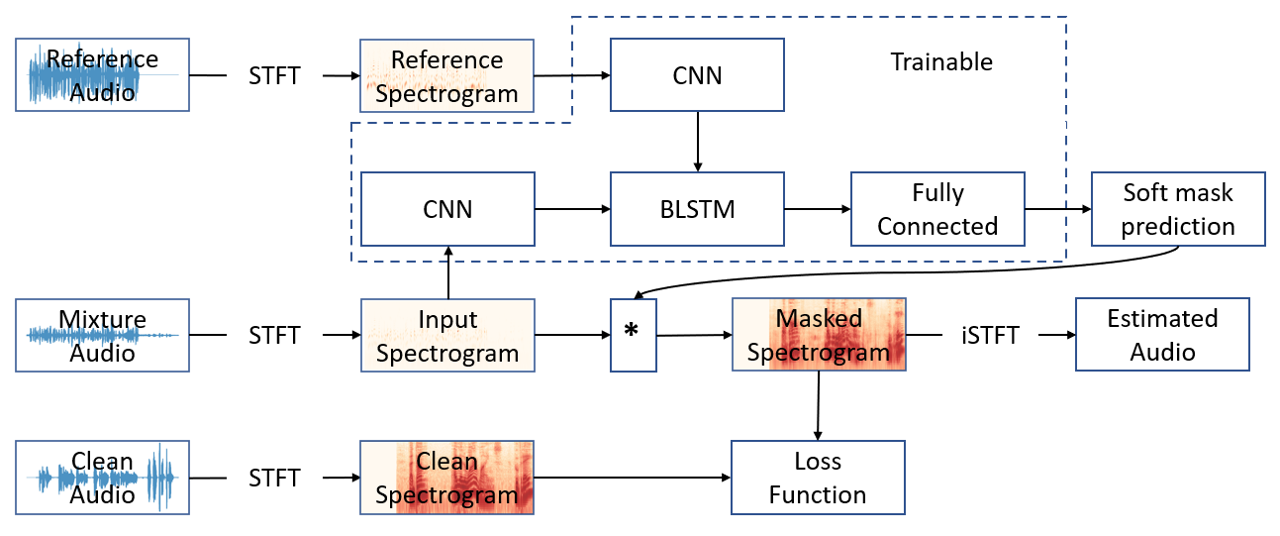}
    \caption{CRNN architecture. }
    \label{fig5}
    \Description{We expect the network to learn from the magnitude spectrograms of reference signal and generate SM to filter the interference sound in the mixture.}
\end{figure*}

We designed a CRNN architecture as shown in Fig.\ref{fig5}. The network predicts a soft mask, which is element-wise multiplied with the mixture magnitude spectrogram to produce an estimated waveform. We directly merge the phase of the noisy audio with the estimated magnitude spectrogram and apply an iSTFT on the result. The network is trained to minimize the difference between the masked magnitude spectrogram and the target magnitude spectrogram computed from the clean audio.
The system consists of two separate convolutional neural networks (CNN), each with eight layers and batch normalization layers, and one bidirectional long-short-time memory (BLSTM) layer, followed by two fully connected (FC) layers. All of these layers have ReLU activations except for the last layer, which has a sigmoid activation. The system takes three inputs for one training step: (1) clean ground truth audio from the target human speaker, (2) noisy audio containing the overlapping speech from the robot and the target speaker, and (3) reference audio generated from the APIs \cite{googletts}.
Because we expect the network to estimate the target speech based on the reference signal instead of the reference speaker identification, we did not adopt the Speaker Encoder in our architecture as \cite{wang2019} and \cite{he2020speakerfilter} did. Instead, we adopted another convolutional neural network to learn from the reference spectrogram, which shares the same hypermeter setting as that for the input spectrogram. 
Parameter values are provided in Table \ref{table:1}.

\begin{table}
\centering
\caption{Parameter setting of the proposed network.}
\label{table:1}
\begin{tabular}{cccccc}
\hline
\multirow{2}{*}{Layer} & \multicolumn{2}{c}{Width} & \multicolumn{2}{c}{Dilation} & \multirow{2}{*}{Filters/Nodes} \\
 & time & freq & time & freq &  \\ \hline
CNN 1 & 1 & 7 & 1 & 1 & 64 \\
CNN 2 & 7 & 1 & 1 & 1 & 64 \\
CNN 3 & 5 & 5 & 1 & 1 & 64 \\
CNN 4 & 5 & 5 & 2 & 1 & 64 \\
CNN 5 & 5 & 5 & 4 & 1 & 64 \\
CNN 6 & 5 & 5 & 8 & 1 & 64 \\
CNN 7 & 5 & 5 & 16 & 1 & 64 \\
CNN 8 & 1 & 1 & 1 & 1 & 8 \\
BLSTM & - & - & - & - & 400 \\
FC 1 & - & - & - & - & 600 \\
FC 2 & - & - & - & - & 601 \\ \hline
\end{tabular}
\end{table}

To train the system, all input audios are truncated with a 5-second length and are converted to single-channel with a sampling rate of 16kHz if necessary. 

\section{Experiments}
\label{section:4}
\subsection{Dataset}
Instead of recording overlapping speech between Pepper and human interlocutors as a reference dataset, we chose to generate a mixed speech signal instead. There are three reasons why we made this decision. First, we intended to create a dataset that resembled the real recorded overlapping speech as much as possible. Second, we adopted an end-to-end supervised learning method to train our proposed network, which requires human labor to label not only the start of robot speech, but also that of human speech in the recorded data. Generating a mixed signal helps considerably reduce this human labor, as the start times can be controlled. Third, TSE models trained on manufactured overlapping datasets have been reported to have good generalizability to real overlapping recordings\cite{zmolikova2023neural}. 

Therefore, we collected three sets of real recorded data for the development and evaluation of the proposed methods: one with the robot playing a sine signal described in Section \ref{section:3} and its ego fan noise, one with a speaker playing human speech, and the other with the robot playing speech generated by the robot's embedded TTS API and Google Dialogflow's TTS API. We used the first to calculate the speaker-microphone frequency response coefficients, the second to determine the human speech power gain when overlapping with robot speech, and the third to generate the overlapping speech data for training and testing.

We used Pepper for all recordings, using one of the four microphones on top of its head with a forward look direction. The audio signals received by the microphone are strongly affected by the fan noise inside its head. The sampling rate is 16 kHz. The corresponding collected data can be found at this link\footnote{10.17605/OSF.IO/V4Y6H}.

\begin{figure}[ht]
    \centering
    \subfigure[Large laboratory room.]
    {
        \includegraphics[width=0.43\linewidth]{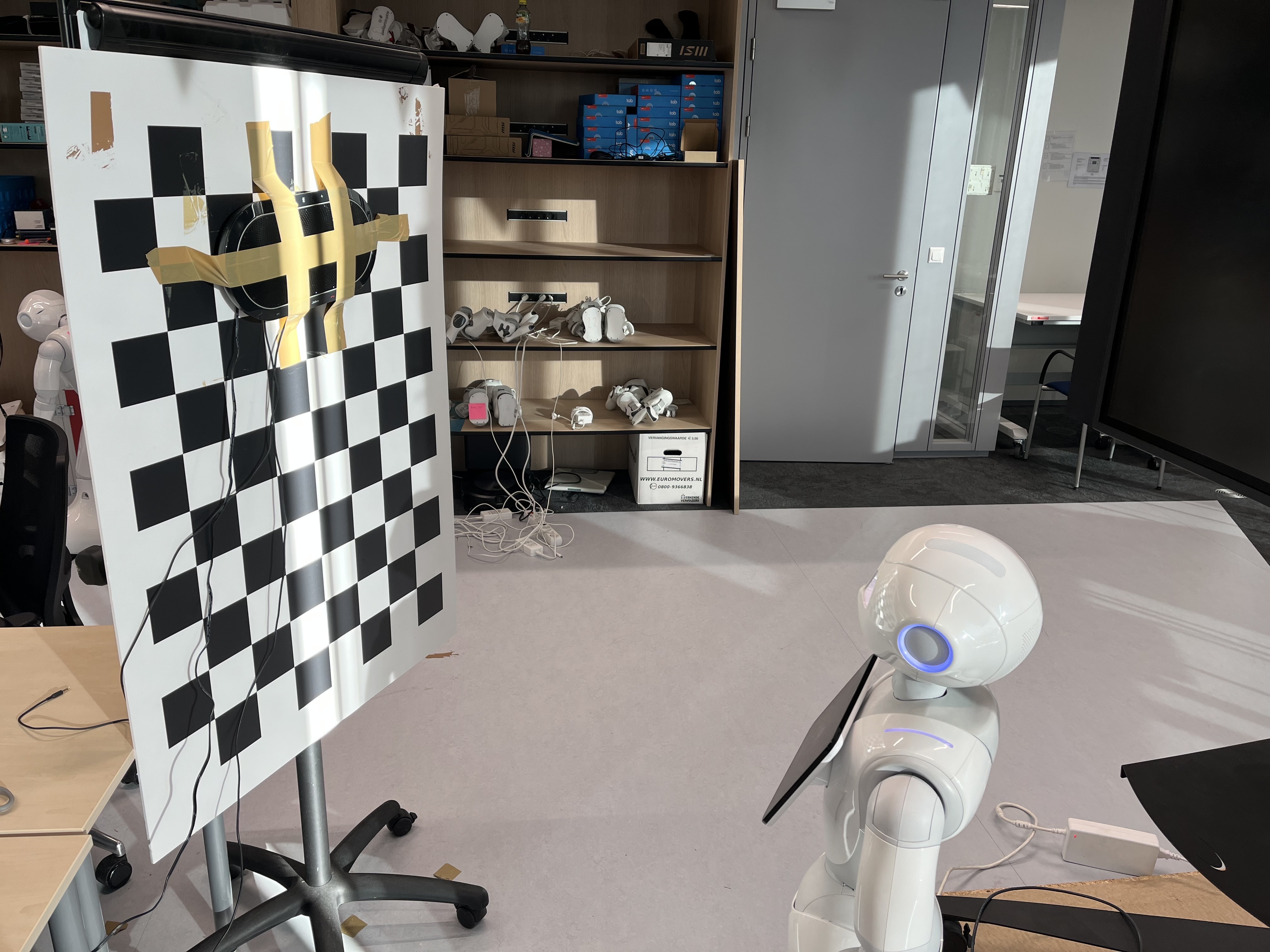}
        \label{fig6:1st_sub}
    }
    \subfigure[Small office room.]
    {
        \includegraphics[width=0.43\linewidth]{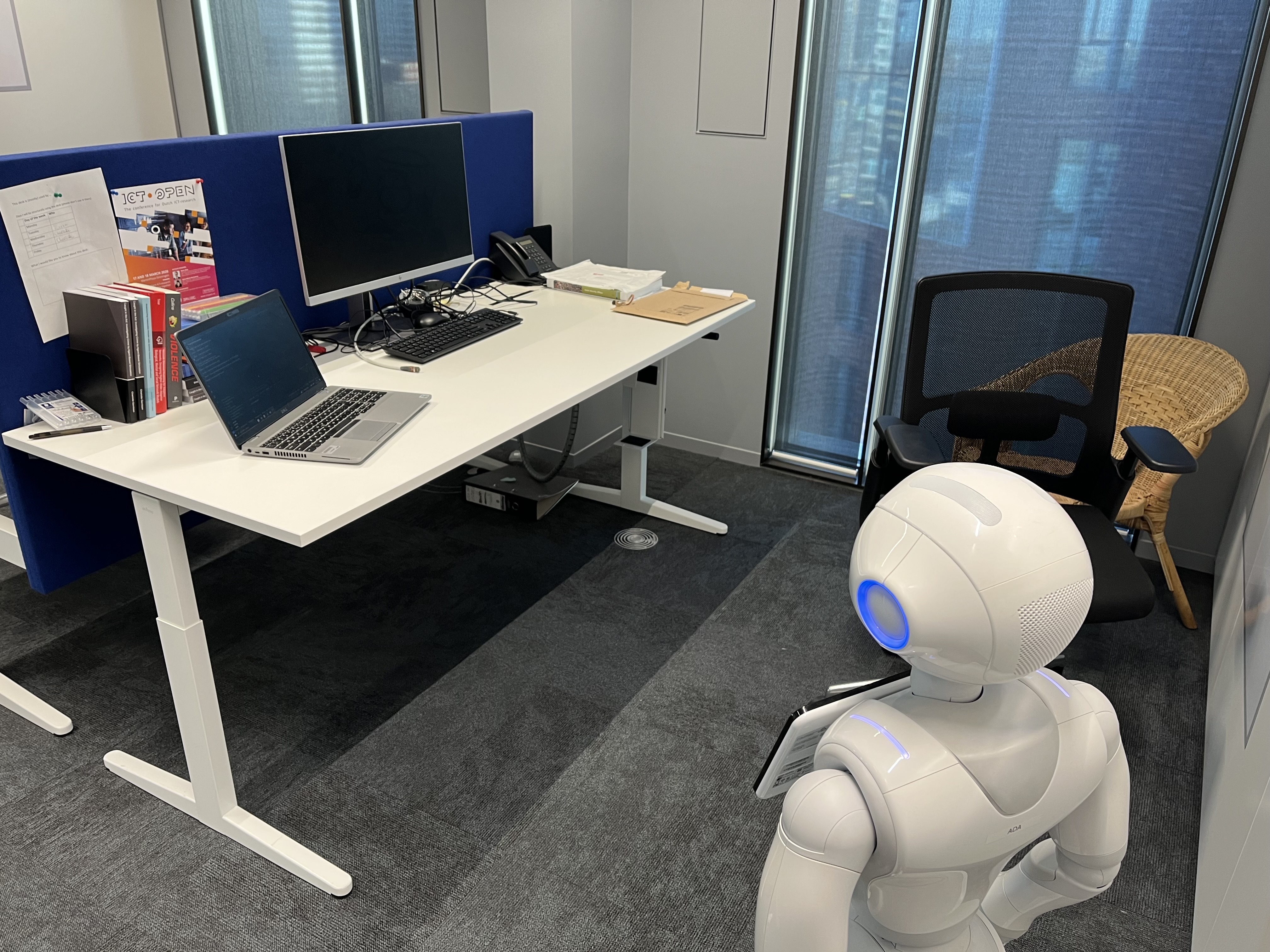}
        \label{fig6:2nd_sub}
    }
    \centering
    \caption{Experiment setup for Data collection with Pepper.}
    \label{fig6}
    \Description{}
\end{figure}

\textbf{Recording Sine Signal and Ego Noise:} We collected these data by recording the sine signal whose frequency sweeps over (0 Hz, 8001 Hz) with a step of 13 Hz. We placed the robot in a large and quiet laboratory room, and programmed it to stand still and look ahead while recording the sound produced by its own speakers at volume 50 (as shown in Fig.\ref{fig6:1st_sub}). We also recorded Pepper's ego fan noise under this condition, without Pepper doing anything but standing still.

\begin{figure}[ht]
    \centering
   
    \subfigure[Recorded human speech at volume 50.]
    {
        \includegraphics[width=0.43\linewidth]{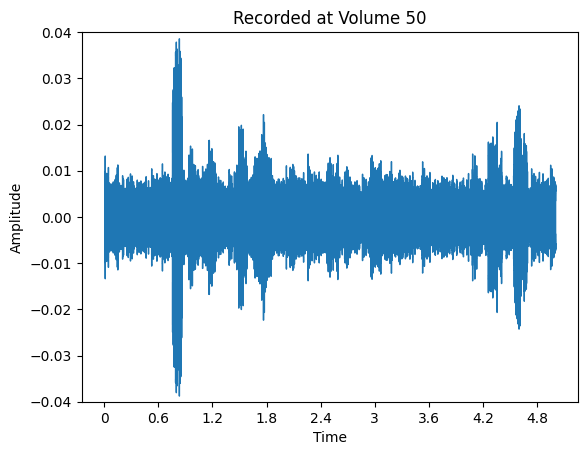}
        \label{fig11a}
    } 
    \subfigure[Manufactured human speech with -25 power gain.]
    {
        \includegraphics[width=0.43\linewidth]{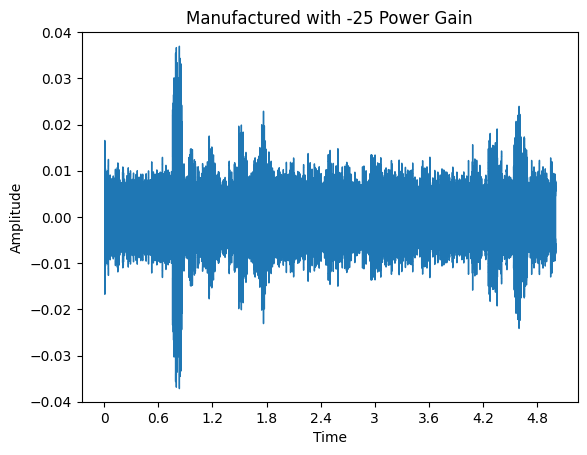}
        \label{fig11b}
    }
    \subfigure[Spectrogram of the recorded signal.]
    {
        \includegraphics[width=0.43\linewidth]{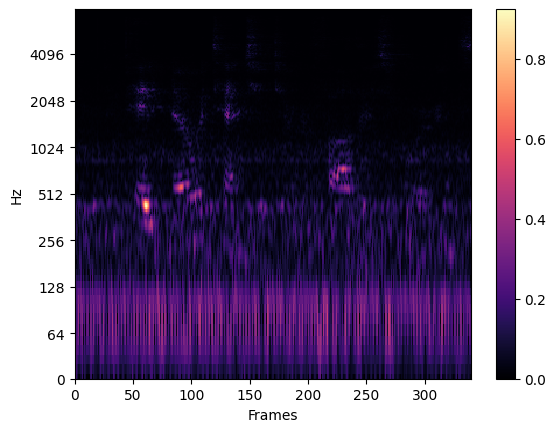}
        \label{fig11c}
    } 
    \subfigure[Spectrogram of the manufactured signal.]
    {
        \includegraphics[width=0.43\linewidth]{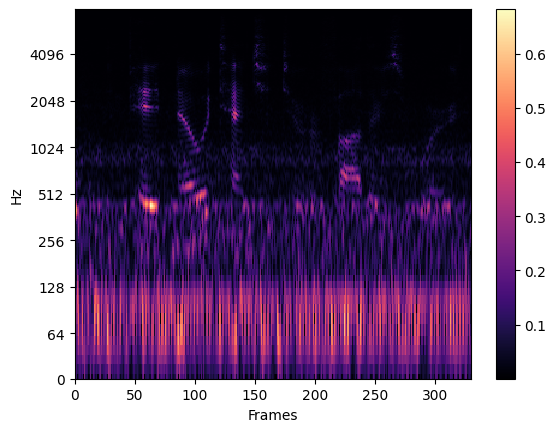}
        \label{fig11d}
    } 
    
    \centering
    \caption{Time domain and frequency domain display of the recorded speech signal and the generated human speech signal played by the speaker.}
    \label{fig11}
    \Description{The average signal-to-noise ratio of manufactured human speech is similar to that of recorded human speech. The spectral values of the speech signal are comparable to those of the fan noise.}
\end{figure}

\textbf{Recording Human Speech:} We collected the data by recording clean speech played on a loudspeaker placed 1 meter from Pepper in the same laboratory room. We programmed Pepper to look at the speaker when the speaker was playing. A total of 1,163 clean speech fragments from the Librispeech corpus\cite{panayotov2015librispeech} were selected. We altered the speaker volume from 10 to 100, and decided that volume 50 was close enough to the common human's volume when interacting with a robot. With this volume, the speech would retain its characteristics in the TF domain from fan noise, as shown in Fig.\ref{fig11c} and Fig. \ref{fig11d}.


\textbf{Recording Robot Speech:} Robot speech was collected by recording Pepper playing the API-generated speech signal, whose length is longer than 5 seconds, from its embedded speakers at volume 50. We used 17 different speaker voices, including Pepper's own embodied one. We created recordings in the laboratory and a small office room (as shown in Fig.\ref{fig6}),  and recorded 7913 audio in total. These 7913 different speech contents were randomly selected from Librispeech \cite{panayotov2015librispeech}. We trimmed the silent parts of the recordings to align with the speech signal in the time domain using cross-correlation. Taking into account reverberation, only the first 5-second segment was selected in each recording. Finally, we used 1800 audio segments recorded in the large laboratory and 6200 in the small office to train and evaluate the proposed methods.

\subsection{Data Generation}
\begin{figure}
    \centering
    \includegraphics[width=\linewidth]
    {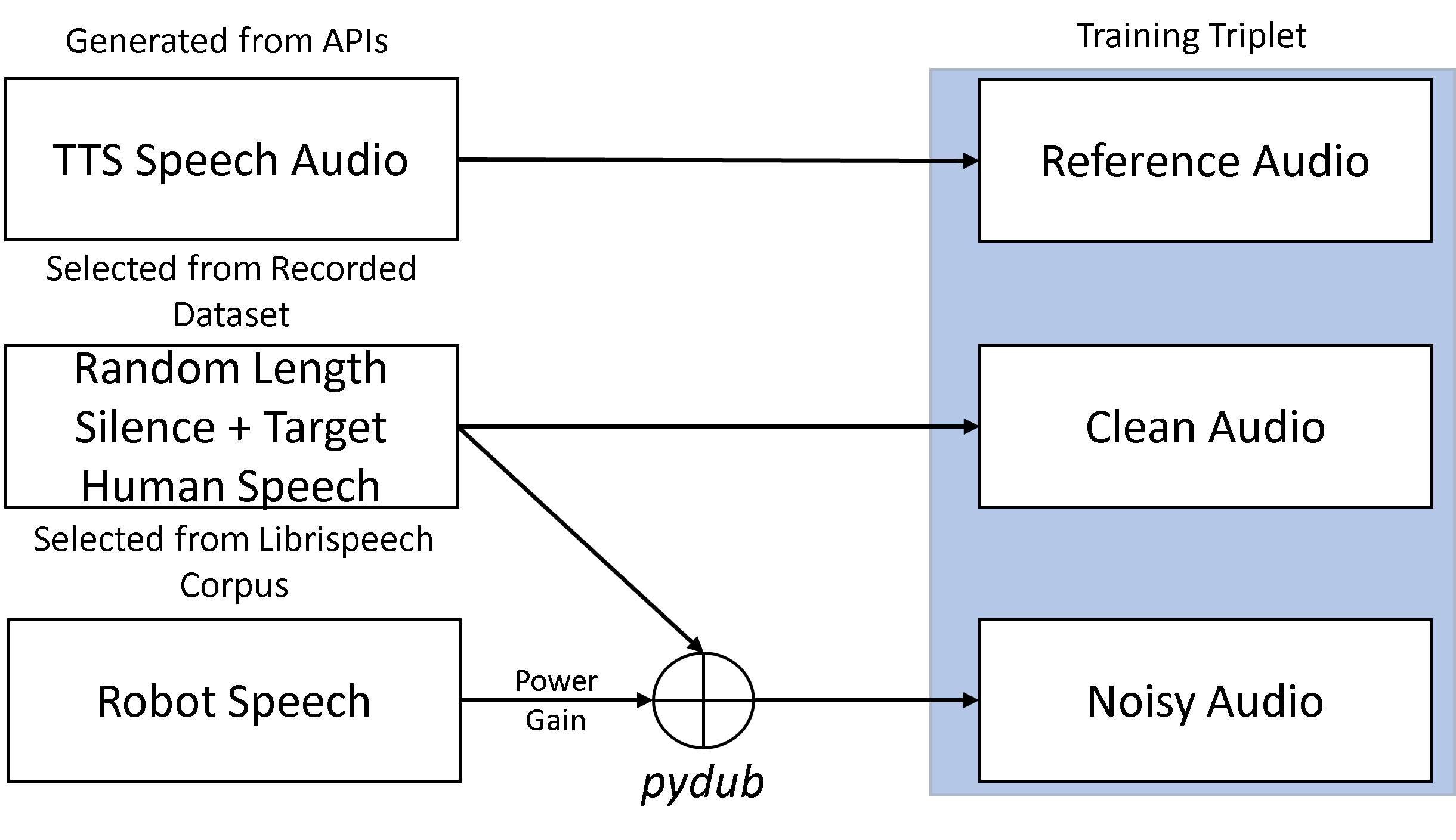}
    \centering
    \caption{Input data processing workflow.}
    \label{fig8}
\end{figure}

We cannot use a "standard" benchmark dataset due to the reasons mentioned in Section \ref{section:1}. Therefore, we use the scheme shown in Fig. \ref{fig8} to obtain the training triplets. The noisy audio is generated by mixing the recorded Pepper speech and the clean speech audio randomly selected from one speaker in the Librispeech dataset. More specifically, it is obtained by directly applying the \textit{overlay} function in the Python \textit{pydub} library. Before \textit{overlay}, a silent segment with random lengths between 0.5 and 1.5 seconds is added before the clean target speech. We set the clean target speech power gain to -25 because the signal-to-fan-noise ratio is close to the real recordings, as shown in Fig.\ref{fig11a} and Fig. \ref{fig11b}. 

Based on the distribution of the recordings in room size, we randomly selected 200 segments recorded in the large laboratory room and 600 in the small office as a validation dataset to evaluate different approaches, while the rest were used for network training.

\subsection{Network Training}
We adopted the power-law compressed reconstruction error \cite{wang2019} as a loss function to train our network and monitored the training process with \textit{Tensorboard} to avoid overfitting. 

\subsection{Baseline Method}
To compare the proposed methods with the state-of-the-art method \cite{wang2019}, we adopted the pre-trained VoiceFilter model provided by Seung-won Park \cite{voicefilter}.  Since this model requires a reference speech signal to learn which voice to filter out, we randomly selected another recorded robot speech file that belongs to the same speaker identification in the same room. This is in line with what was done in \cite{wang2019}.

\subsection{Evaluation Metrics}
To evaluate the performance of the three different proposed methods, we use three metrics: the speech recognition Word Error Rate (WER), the Scale-Invariant Signal-to-Distortion Ratio (SI-SDR) between the estimated signal and the target speech, and the computing time.

\subsubsection{Word Error Rate}
As mentioned in Section \ref{section:1}, the main goal of our system is to improve speech recognition when human speech overlaps with robot speech. We chose the state-of-the-art open-source Whisper \cite{radford2022whisper} ASR system for WER evaluation. Because we truncated human speech at the point where robot speech ends in each fragment, we cannot use ground-truth transcription. Instead, we used the transcription of clean human speech truncated at the same point as the ground truth to calculate the WER.

We calculated the WER value after processing the overlapping audio using each of the methods. As a reference, we also calculated the WER on the overlapping audio without any processing performed. A good filtering system should be able to reduce the WER significantly, which means that this system is improving human speech recognition when a robot is actively speaking itself.

\subsubsection{Scale-Invariant Signal-to-Distortion Ratio}
The SI-SDR is a 
common metric to evaluate single-channel speech separation systems \cite{le2019sdr}. It is an energy ratio, expressed in dB, between the orthogonal projection of the estimated signal on the spanned line of the target speech signal. A higher value indicates better performance. 

\subsubsection{Computing Time}
The computing time required to process the input signal is crucial for the real-life application of a TSE system during HRI. Therefore, we present the computing time for each proposed method to process a noisy speech mixture with a 5 second length. All calculations are done on a local desktop with an Intel(R) Core(TM) i9-9900K CPU and a NVIDIA GeForce RTX 2070 SUPER GPU for network training acceleration.

\section{Results and analysis}
\label{section:5}
In Table \ref{table2}, we present the results of the proposed methods in different rooms compared to the original overlapping files and the baseline model.

\begin{table*}[]
\centering
\caption{TSE model performance on the test set. The laboratory and small room differ in the extent to which the recorded robot speech is interfered by its reverberation.}
\label{table2}
\begin{tabular}{ccclcclcclcclc}
\hline
Scenario & \multicolumn{6}{c}{Big Laboratoy} & \multicolumn{6}{c}{Small Office} &  \\
\multirow{2}{*}{Metrics} & \multicolumn{3}{c}{WER /\%} & \multicolumn{3}{c}{SI-SDR /dB} & \multicolumn{3}{c}{WER /\%} & \multicolumn{3}{c}{SI-SDR /dB} & Computing \\
 & Mean & Median & Std & Mean & Median & Std & Mean & Median & Std & Mean & Median & Std & Time /s \\ \hline
Unfiltered & 138.5 & 130.2 & 50.5 & -22.00 & -21.6 & 3.28 & 138.0 & 120.2 & 90.5 & -26.3 & -26.0 & 4.29 & - \\
Baseline & 130.6 & 120.6 & 51.0 & -19.17 & -18.79 & 3.45 & 130.5 & 112.7 & 87.6 & -24.18 & -23.57 & 4.73 & 1.032 \\
SS before post-filtering & \textbf{47.9} & \textbf{38.0} & 38.1 & -11.9 & -10.1 & 6.18 & 102.9 & 93.3 & 88.9 & -25.4 & -25.2 & 4.99 & \textbf{0.854} \\
SS after post-filtering & 69.1 & 70.1 & \textbf{19.4} & -37.0 & -36.0 & 12.13 & 97.3 & 86.5 & 75.3 & -42.5 & -41.0 & 10.50 & 1.565 \\
CRNN & 63.4 & 66.7 & 31.3 & \textbf{-2.9} & \textbf{-2.5} & \textbf{3.50} & \textbf{68.8} & \textbf{76.7} & \textbf{31.5} & \textbf{-4.1} & \textbf{-4.0} & \textbf{3.88} & 1.060 \\ \hline
\end{tabular}
\end{table*}
\subsection{Results}
We compare the mean, median, and standard deviation values of the estimated speech WER and SI-SDR between the proposed methods and the baseline methods, with the original unfiltered overlapping speech data as a reference. We can observe that the baseline model does not filter out robot speech in most of the files. In fact, the WER and SI-SDR are close to the original unfiltered data. In contrast, the proposed signal processing-based pipeline without post-filtering has the best WER under the weak reverberation condition. There is a significant gap between the results before and after post-filtering in each condition. 

However, when overlapping speech is strongly polluted by the reverberation of robot speech, this method does not significantly improve the ASR result. In comparison, the proposed CRNN-based method shows robustness for each condition with only a slightly higher average WER in comparison to the low-reverberation performance, although the WERs are still greater than 50\%.  The estimated speech from CRNN has the best SI-SDR results, although they are still less than 0.

The computing time required to process 5-second-long audio by the signal processing-based pipeline without post-filtering is the shortest, as low as 854 
milliseconds. However, the time required by other methods is close to that. The proposed CRNN consumes an acceptable 28 milliseconds more than the baseline network. This shows promise for application of the proposed methods in real-life HRI.

\subsection{Discussion}
On the basis of the analysis of the results, we find that the reverberation of the room limits our proposed signal processing-based architecture to practical application in real life. This is because the proposed signal processing-based pipeline is able to filter out the robot's ego speech, but has no impact on the reverberation. 
There are two factors that contribute to this. First, the residual of the robot speech still has a relatively larger power compared to the target speech after filtering. Second, to filter out robot speech and emphasize target speech in the TF domain, the parameters $\alpha$ and $\beta$ in Equations \ref{eq4} and \ref{eq9} result in oversubtraction (e.g. parts of the target speech are also removed), which will further result in distortion in the estimated signal. This distortion is reinforced by the pre-trained post-filtering network. This also explains why SI-SDR after post-filtering drops significantly in each condition. When the reverberation is low, the target human speech will possess a relatively greater power in the estimated speech, and the ASR system will translate this instead of reverberation. However, when the reverberation is strong and has greater power than the target speech, the ASR system cannot recognize the target speech and instead translates the reverberation. 
We need to emphasize that it is impractical to use a simple filter, such as the Wiener filter, to filter out the reverberation of robot speech, because the filter will regard low-energy human speech as noise and eliminate the target speech in the estimated signal. Furthermore, the amplify coefficient $\beta$ in Equation \ref{eq9} cannot be set too high to prevent distortion.

A possible solution to improve the ASR result of the signal processing-based method is to perform mask filtering not only on the magnitude of the overlapping speech spectrogram but also on the phase. For example, Donald S Williamson et al. \cite{williamson2015complex} proposed to perform complex ratio masking for monaural speech separation and got great performance in perceptual evaluation of speech quality. A second potential solution is to use the adaptive step size method \cite{nakajima2009blind} to generate the SM. For example, we can time-shift the SM and perform spectral subtraction on the estimated speech signal iteratively. However, this requires more computational time or an estimate of the number of iterations.

\begin{figure}
    \centering
    \includegraphics[width=\linewidth]
    {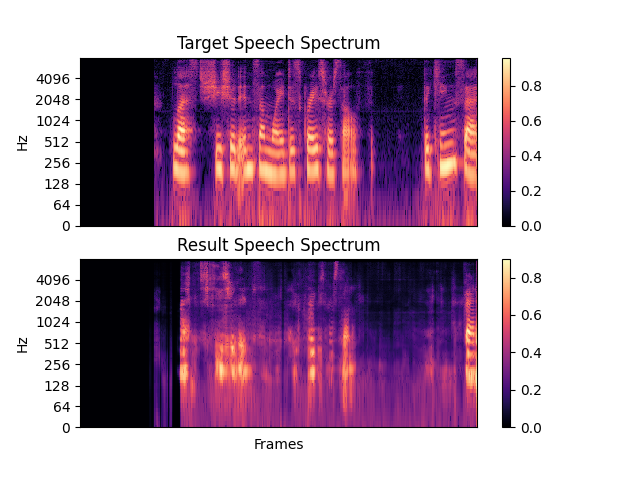}
    \caption{The spectrogram of the CRNN estimated signal and the target speech signal. }
    \label{fig10}
    \Description{Some spectral values in high frequency bins are discarded.}
\end{figure}

Another reason why the ASR results of each estimated signal are significantly different is that the weak or low pitched target speech in our fabricated segments posed a problem for both the signal processing pipeline and the CRNN.  This is because some values in the essential frequency bins are discarded in the spectrogram of the estimated speech signal, as shown in Fig.\ref{fig10}. These discarded characteristics in high-frequency bins in the Time-Frequency domain are important for the ASR system because they contain more details on the speakers \cite{deng2016deep}. 

For the CRNN, a different problem is at play. In our setup, we included a batch normalization layer, in accordance with \cite{wang2019}. The role of this layer is to normalize the spectrum values based on all segments in a given batch. However, robot speech often possesses much more power compared to human speech if they overlay at the same point on the spectrogram. As a result, we found that the normalization procedure reduced the value of human speech to a value that is too small to be learned during training. It is therefore not practical to adopt a batch normalization layer in the TSE models when the interfering speech is significantly more powerful than the target speech. Another reason why the CRNN result is unsatisfactory is that the training data set is as small as 7916 compared to the baseline network, which used 100,000 triplets for training. 

Another takeaway from the result is that the WER does not directly relate to the SI-SDR, which means that state-of-the-art ASR systems are tolerant to some distortion in human speech. It demonstrates the necessity to report not only the distortion of the restored speech signal but also the WER of the restored speech contents, which is the key concern in HRI research.


We compared the results of the signal processing pipeline and CRNN with a baseline model based on speaker identification, which yielded considerably worse performance. This is surprising, given that \cite{wang2019} claimed that their model using speaker identification as reference showed strong robustness when the interference signal had more power than the target speech. However, in our focused task, it was not possible to estimate the target speech when the robot speech possesses greater power than the target human speech. In fact, their model failed to filter out robot speech interference for most of the files.

\section{Conclusion and Future Work}
\label{section:6}
In this paper, we designed and evaluated two different architectures that focus on filtering the robot speech signal from the robot received signal and improving the ability to recognize human speech during interaction when the humanoid robot and the human are both actively speaking. We demonstrated the effectiveness of these proposed methods on a manufactured dataset of real-recorded human and robot speech. We found that the proposed signal processing-based pipeline without post-filtering was able to improve the ASR ability when the reverberation of the room is weak in real time and the target speech is high pitched or at a relatively high volume. The proposed CRNN also showed good robustness to each condition, but the performance was still not satisfactory. 

In terms of future work, we will look for more possible methods to improve performance. For the signal processing-based pipeline, a dereverberation speech mask should be designed to filter out the reverberation of robot speech. For the neural network-based architecture, we need to construct a larger dataset for training. Furthermore, instead of applying iSTFT to recover the estimated signal, a decoder network should be adopted. Furthermore, we will train the network jointly with an ASR tokenizer to further increase the improvement in WER. And last, we will also apply the proposed methods to real-life HRI to evaluate the method's effectiveness in identifying the human subjects' interruption and backchanneling.

\bibliographystyle{ACM-Reference-Format}
\bibliography{mylib}

\end{document}